\pdfoutput=1

\documentclass[11pt]{article}

\usepackage[final]{acl}

\usepackage{times}
\usepackage{latexsym}
\usepackage{amsmath}
\usepackage{multirow}
\usepackage{booktabs}
\usepackage{amssymb}
\usepackage{arydshln}
\usepackage{graphicx}
\usepackage[T1]{fontenc}

\usepackage[utf8]{inputenc}

\usepackage{microtype}

\usepackage{inconsolata}

\usepackage{graphicx}

\title{Synergizing LLMs with Global Label Propagation for Multimodal Fake News Detection}

\author{
 \textbf{Shuguo Hu\textsuperscript{1}},
 \textbf{Jun Hu\textsuperscript{2}\thanks{Corresponding author: Jun Hu.}},
 \textbf{Huaiwen Zhang\textsuperscript{1}}
\\
 \textsuperscript{1}Inner Mongolia University,
 \textsuperscript{2}National University of Singapore
\\
shuguo.hu@mail.imu.edu.cn, jun.hu@nus.edu.sg, huaiwen.zhang@imu.edu.cn
}

\begin{document}
\maketitle
\begin{abstract}
Large Language Models (LLMs) can assist multimodal fake news detection by predicting pseudo labels.
However, LLM-generated pseudo labels alone demonstrate poor performance compared to traditional detection methods, making their effective integration non-trivial.
In this paper, we propose Global Label Propagation Network with LLM-based Pseudo Labeling (GLPN-LLM) for multimodal fake news detection, which integrates LLM capabilities via label propagation techniques. 
The global label propagation can utilize LLM-generated pseudo labels, enhancing prediction accuracy by propagating label information among all samples.
For label propagation, a mask-based mechanism is designed to prevent label leakage during training by ensuring that training nodes do not propagate their own labels back to themselves.
Experimental results on benchmark datasets show that by synergizing LLMs with label propagation, our model achieves superior performance over state-of-the-art baselines. 
Our code is available online~\footnote{\url{https://github.com/TSCenter/GLPN-LLM}}.

\end{abstract}

\begin{figure}[t]
	\centering
	\includegraphics[width=1\linewidth]{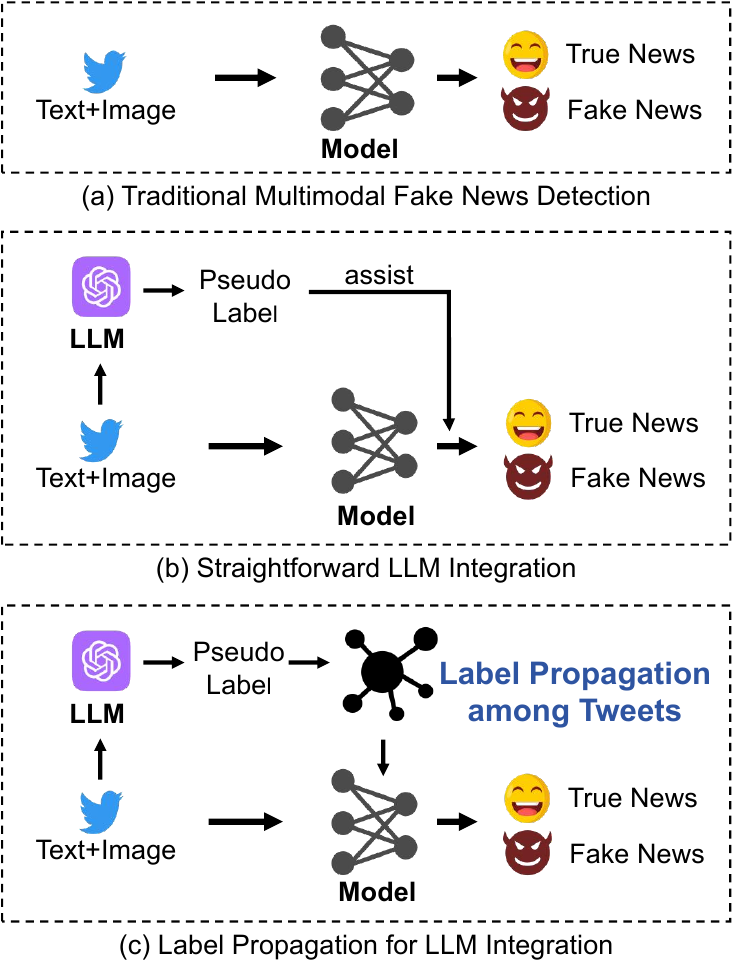}
	\caption{
    Illustrations of different methods.
 }
 \label{fig:figure1}
\end{figure}

\section{Introduction}

Detecting and mitigating the spread of multimodal fake news is a critical task for safeguarding the authenticity of information in the digital age~\cite{zhang2020overview}, as shown in Figure~\ref{fig:figure1}(a).
In recent years, the rapid growth of social media platforms has significantly accelerated the spread of misinformation, underscoring the urgent need for effective detection techniques~\cite{shu2017fake, zhou2020survey, shu2019beyond, perez2017automatic, zhou2019fake}.
Large Language Models such as GPT-4~\cite{achiam2023gpt, brown2020language} have demonstrated strong capabilities in language understanding and reasoning tasks~\cite{DBLP:conf/acl/XiongPKF24,DBLP:conf/icml/0008YSX24,chu2024towards,DBLP:conf/acl/BaiWZGLS24}, making them promising tools for enhancing fake news detection systems~\cite{wu2024fake, sun2024exploring, hu2024bad, su2023adapting}.

A straightforward approach to leveraging LLMs for multimodal fake news detection involves directly combining predictions from existing models with LLM outputs, as illustrated in Figure~\ref{fig:figure1}(b).
However, LLM-generated pseudo labels may significantly underperform compared to existing multimodal fake news detection models (see Table~\ref{tab:benchmark_maskllm}), indicating that this direct combination approach requires further refinement.
Therefore, it is essential to explore more effective methods for integrating LLM capabilities into fake news detection tasks.

To address these limitations, we propose a novel framework that integrates LLM-generated pseudo labels via Label Propagation (LP)~\cite{zhur2002learning} techniques, as shown in Figure~\ref{fig:figure1}(c).
LP enhances classification performance by propagating labels or pseudo labels between samples~\cite{zhur2002learning, iscen2019label, fcn-lp}. 
Importantly, LP can remain effective even when pseudo label accuracy is moderate~\cite{DBLP:journals/pr/SunHGCLY25}, making it well-suited for incorporating LLM-generated pseudo labels in fake news detection, where individual LLM predictions may be imperfect.

Our framework, Global Label Propagation Network with LLM-based Pseudo Labeling (GLPN-LLM), introduces a mask-based global label propagation module that works alongside an LLM-based pseudo label generation module.
The global label propagation module can utilize LLM-generated pseudo labels, enhancing prediction accuracy by propagating label information among all samples.
For label propagation, a mask-based mechanism is designed to prevent label leakage during training by ensuring that training nodes do not propagate their own labels back to themselves.
Experimental results on benchmark datasets show that by synergizing LLMs with label propagation, our model achieves superior performance over state-of-the-art baselines, demonstrating its effectiveness for fake news detection.

In summary, our contributions are threefold:
\begin{itemize}
\item We propose Global Label Propagation Network with LLM-based Pseudo Labeling (GLPN-LLM), a novel multimodal fake news detection framework that integrates LLM capabilities via label propagation techniques.
\item We introduce a mask-based global label propagation mechanism that prevents label leakage during training while effectively propagating label information across all samples.
\item We conduct experiments on three benchmark datasets, demonstrating that our framework achieves superior performance compared to state-of-the-art baselines with significant improvements in accuracy and F1 scores.
\end{itemize}

\begin{figure*}
    \centering
     \includegraphics[width=1\linewidth]{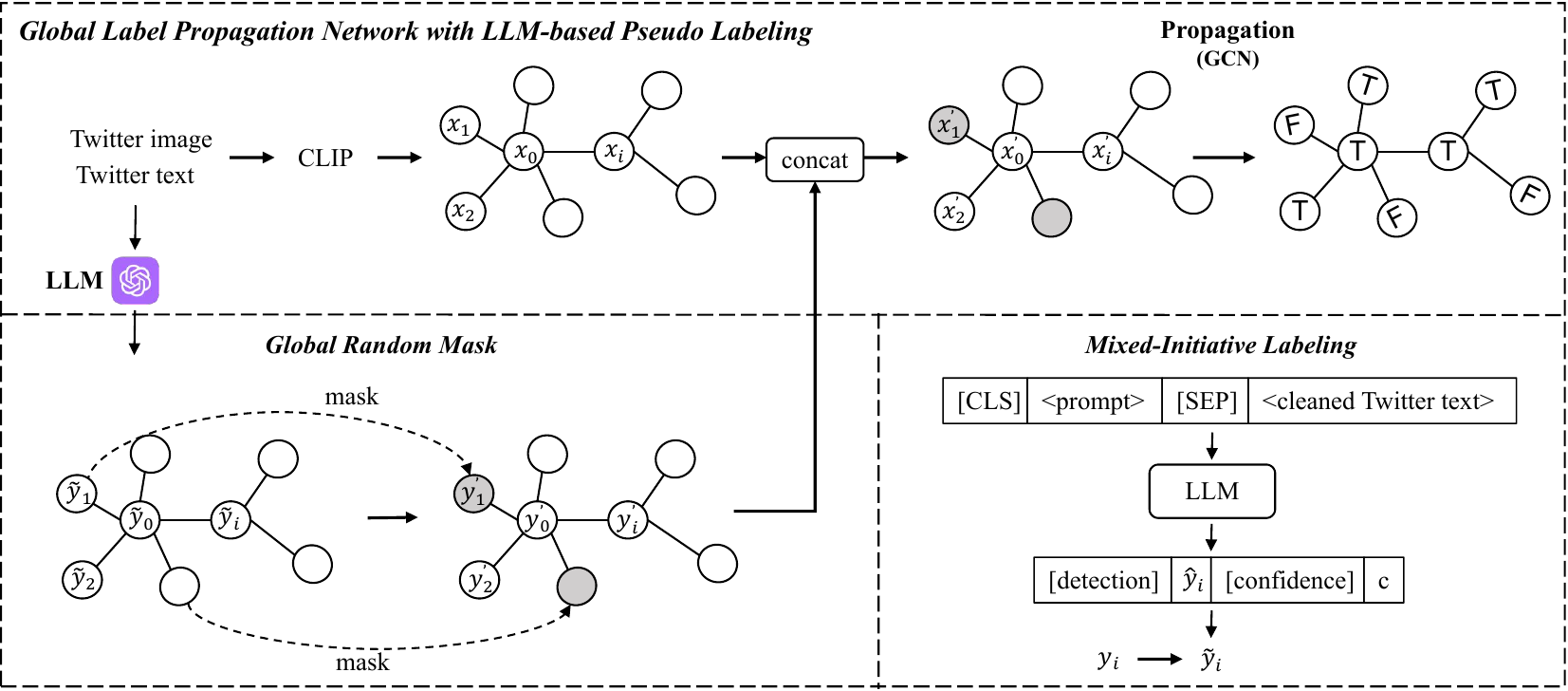}
    \caption{
        Overview of the GLPN-LLM framework for fake news detection. The framework synergizes LLM-generated pseudo labels with a global label propagation mechanism, leveraging multimodal features.
    }
    \label{fig:figure2}
\end{figure*}

\section{Related work}
\label{sec:relatedwork}

In this section, we review related work in multimodal fake news detection and label propagation techniques.

\subsection{Multimodal Fake News Detection} 
Early fake news detection methods primarily focused on text-based classification~\cite{shu2017fake, wang2018eann}. 
Recent work in fake news detection has also benefited from multimodal learning~\cite{jin2017multimodal}, which has achieved success in many applications~\cite{gao2025learning,DBLP:conf/aaai/0016HHW25,DBLP:conf/iccv/ZhangGYLH23,DBLP:journals/pami/YangXHX24,DBLP:journals/tkde/FangZHWX23}.
Recent studies show growing attention on multimodal representations~\cite{yang2021rumor, spotfake} and various multimodal methods have been proposed, such as adversarial training for modality-invariant feature learning~\cite{wang2018eann}, multimodal attention mechanisms~\cite{hmcan}, and multimodal graph-based approaches~\cite{DBLP:conf/mir/WangQHFX20,fcn-lp}.

\subsection{Label Propagation} 

Label Propagation (LP)~\cite{zhur2002learning} spreads label information across a graph to predict unlabeled nodes, assuming connected nodes may share labels~\cite{zhu2003semi}. 
It has been extended to improve performance, such as kernelized LP for non-linear relationships~\cite{zhou2003learning} and post-processing approaches that correct predictions through error correlation~\cite{DBLP:conf/iclr/HuangHSLB21}.
Some recent work~\cite{DBLP:conf/kdd/ZhangYS0OLT0022,DBLP:conf/aaai/YangYPYF23} combines LP with Graph Neural Networks (GNNs)~\cite{kipf2016semi,DBLP:conf/iclr/VelickovicCCRLB18,DBLP:conf/www/MaoLLS23,DBLP:journals/tkde/HuHH24}---which show promising performance for social media applications~\cite{DBLP:conf/icde/LiangZSYJT024, DBLP:journals/tkde/HuHQFX24, DBLP:conf/mm/HuQFWZZX21, DBLP:conf/www/ZhangLY0QQX24, sun2023neighborhood, DBLP:journals/tkde/SangWZZW25, DBLP:journals/tois/SangLZZY25, qiao2024deep, DBLP:conf/www/0023H25}---achieving encouraging results.

\section{Method}

In this section, we introduce our Global Label Propagation Network with LLM-based Pseudo Labeling (GLPN-LLM) framework.

\subsection{Global Label Propagation Network with LLM-based Pseudo Labeling}
LLM-generated pseudo labels underperform compared to existing multimodal fake news detection models.
This underperformance makes their effective integration into detection systems a significant challenge. 
Therefore, as simpler direct combination methods require further refinement, we explore more advanced strategies to fully leverage the potential of LLMs. 
To address these limitations, we propose the Global Label Propagation Network with LLM-based Pseudo Labeling (GLPN-LLM) framework.
This novel framework enables comprehensive label propagation across the entire graph and leverages Large Language Models to generate pseudo labels for the test set. By integrating these components, GLPN-LLM ensures full data utilization and improves label alignment between text-image representations and their corresponding labels, thereby significantly enhancing the effectiveness of fake news detection.

\subsection{Multimodal Feature Extraction}

To effectively capture the multimodal characteristics of news items, we employ CLIP~\cite{clip} for feature extraction. CLIP is a powerful model that jointly learns visual and textual representations by aligning them in a shared embedding space. Given a news item comprising an image and its corresponding text, we utilize CLIP’s dual encoders to generate high-dimensional feature vectors for both modalities.

Specifically, the image encoder produces the visual feature vector \( \mathbf{v_i} \in \mathbb{R}^{d_v} \), while the text encoder yields the textual feature vector \( \mathbf{t_i} \in \mathbb{R}^{d_t} \).
These feature vectors are then concatenated to form a unified representation \( \mathbf{x_i} \in \mathbb{R}^{d_t + d_v} \), where:
\begin{equation}
    \mathbf{x_i} = \mathbf{t_i} \oplus \mathbf{v_i}
\end{equation}
Here, \( \oplus \) denotes the concatenation operation.
By leveraging CLIP’s robust feature extraction capabilities, our framework generates unified feature vectors that effectively integrate textual (semantic) and visual information from each news item. Ensuring these modalities are well-represented and aligned is crucial for enhancing the overall performance of fake news detection.

\subsection{Cross-Modal Graph Construction}

We construct a cross-modal graph following the graph construction method proposed in FCN-LP~\cite{fcn-lp}. Each node in the graph represents a distinct news item, characterized by a unified feature vector. Edges between nodes are established based on multiple similarity metrics to encapsulate both intra- and inter-modal relationships. These similarity measures include: \textit{1)} \textit{concatenated feature similarity}: This integrates both textual and visual embeddings by calculating the cosine similarity between the concatenated feature vectors of two news items.
\textit{2)} \textit{image-to-text similarity}: This measures the semantic similarity between the image feature of one news item and the text feature of another.
\textit{3)} \textit{text-to-image similarity}: This assesses the similarity from text to image across different news items.
\textit{4)} \textit{image-to-image similarity}: This captures the similarity within the same modality by comparing image features of different news items.
\textit{5)} \textit{text-to-text similarity}: This evaluates the similarity between text features of different news items.
An edge is created between two nodes \( i \) and \( j \) if any of the aforementioned similarity scores exceed a predefined threshold (\( \theta = 0.95 \)). This threshold ensures that only strongly related news items are connected.

By constructing the graph using these comprehensive similarity measures, we ensure that label information can be effectively propagated  across multimodally related news items. 
This cross-modal graph structure leverages the full spectrum of available data, enabling robust and accurate alignment between text-image representations and their corresponding labels. 
Consequently, the Global Label Propagation Network can maximize data utilization and enhance the overall performance.

\subsection{Label Integration into Node Features}

To facilitate effective label propagation across the entire dataset, it is essential to integrate label information directly into the node features within the cross-modal graph. This involves incorporating both the ground truth labels available for training data and the pseudo labels generated for unlabeled or test data. We introduce the Label Integration Module to achieve this, which seamlessly embeds these varied label types into node feature representations, thereby enhancing the semantic alignment between labels and multimodal data.

The core of this module is the construction of an augmented feature vector for each node \( i \). 
Specifically, the label-based feature \( \mathbf{y_i'} \), whose composition is detailed in Section~\ref{global-random-mask}, is concatenated with the original node features \( \mathbf{x_i} \):
\begin{equation}
   \mathbf{x_i'} = \mathbf{x_i} \oplus \mathbf{y_i'}
\label{eq:concat_x_y}
\end{equation}
This integration ensures that both feature and label information are jointly represented within each node, enabling the model to leverage label semantics during the propagation process.

\noindent \textbf{Mixed-Initiative Labeling} To further enhance the label propagation process and ensure the inclusion of high-confidence pseudo labels, we introduce the Mixed-Initiative Labeling approach. This method leverages a pre-trained LLM to generate reliable pseudo labels for unlabeled data, thereby augmenting the label information available for propagation within the graph.

The Mixed-Initiative Labeling process begins with constructing a structured prompt that incorporates both the context and the specific task requirements. The prompt is formulated as 
\begin{equation}
\small
X = \texttt{[cls]} \; \texttt{<prompt>} \; \texttt{[SEP]} \; \texttt{<cleaned Twitter text>}
\end{equation}
\begin{equation}
\small
Y = \texttt{[detection]}, \; \mathbf{\hat{y}}, \; \texttt{[confidence]}, \; c
\end{equation}
The \texttt{<prompt>} provides the necessary context for the LLM to perform the detection task, while the \texttt{<cleaned Twitter text>} represents the preprocessed content of the news item. Detailed examples of the prompts used are provided in Table~\ref{table3}.

Upon receiving the structured input, the LLM processes the prompt and generates two key outputs:
\textit{1) Detection Label} ($\mathbf{\hat{y}}$) indicates the authenticity of the news item, categorizing it as either true or fake.
\textit{2) Confidence Score} ($c$) reflects the LLM's confidence in its prediction, quantifying the probability associated with the generated label.

To ensure the reliability of the propagated labels, we employ a confidence-based filtering mechanism. Specifically, pseudo labels are selected for integration into the graph based on their associated confidence scores (\( c \)). 
Once selected, the pseudo labels (\( \mathbf{\hat{y}} \)) are converted into one-hot encoded vectors and integrated into the graph's node features.
This integration is performed as follows:
\begin{equation}
\scalebox{0.85}{
\(
\mathbf{\tilde{y}_i} = 
\begin{cases} 
\mathbf{y_i} & \text{if node } i \text{ is truly labeled}, \\[6pt]
\mathbf{\hat{y}_i} & \text{if unlabeled node } i \text{ is high-confidence}, \\[6pt]
\mathbf{0} & \text{otherwise}.
\end{cases}
\)
}
\end{equation}
where nodes without high-confidence pseudo labels retain zero vectors as label embeddings, ensuring only reliable label information is propagated.

\subsection{Global Random Mask for Optimization}
\label{global-random-mask}

During inference, we set the label-based features $\mathbf{y_i' = \tilde{y}_i}$.
However, during training, $\mathbf{y_i'}$ is obtained via a Global Random Mask (GRM) to prevent label leakage.
Without a mechanism like the Global Random Mask, if a node's label is included in its input features (due to label integration), the model might learn to predict this label trivially during training, without sufficiently leveraging the graph structure or original content features. This is a form of label leakage.
Using these propagated labels as part of the node input features can inadvertently leak label information into the training process, resulting in biased training outcomes.
GRM addresses this issue by randomly masking the label information of a subset of nodes and computing the loss only for these nodes during each training iteration, where label masking prevents label leakage.

The GRM operates by selectively masking the label embeddings of a randomly chosen subset of nodes in the unified graph during each training epoch. Specifically, given a mask ratio \( \rho \) (e.g., \( \rho = 0.3 \)), a proportion of \( \rho \times N \) nodes are randomly selected, where \( N \) is the total number of nodes in the graph. For each selected node \( i \), its label embedding \( \mathbf{\tilde{y}_i} \) is replaced with a zero vector:
\begin{equation}
\mathbf{y_i'} = \mathbf{\tilde{y}_i} \cdot m_i
\end{equation}
where $m_i$ is a binary mask scalar defined as:
\begin{equation}
\small
m_i = 
\begin{cases}
0 & \text{if node } i \text{ is selected for masking}, \\
1 & \text{otherwise}.
\end{cases}
\end{equation}
Note that this masking operation only affects the label embeddings, leaving the original feature vectors $\mathbf{x_i}$ unchanged. 
This masking prevents label leakage during training by ensuring that training nodes do not propagate their own labels back to themselves, thus preventing the model from relying on this information for classification.

Given $\mathbf{y_i'}$, we obtain $\mathbf{x_i'}$ using Equation~\ref{eq:concat_x_y} as each node's feature vector, then apply a GCN~\cite{kipf2016semi} on the graph to predict labels for each node.
Since $\mathbf{x_i'}$ contains both multimodal content features and label information, both types of information are propagated through the graph by the GCN for fake news detection.
The GCN output is optimized using cross-entropy loss with the Adam optimizer~\cite{adam}.

\begin{table*}[h]
    \centering
    \renewcommand{\arraystretch}{1.4}
    \setlength{\tabcolsep}{2pt}

    \resizebox{1\linewidth}{!}{
    \begin{tabular}{lcccccccccccc}
        \toprule
         \multirow{2}{*}[-2pt]{{Methods}} & \multicolumn{4}{c}{Twitter} & \multicolumn{4}{c}{PHEME} & 
         \multicolumn{4}{c}{Weibo}
         \\
         \cmidrule(lr){2-5}
         \cmidrule(lr){6-9}
         \cmidrule(lr){10-13}
         & \small{Accuracy $\uparrow$} & \small{Precision $\uparrow$} & \small{Recall $\uparrow$} &  
         \small{F1 $\uparrow$} & 
         \small{Accuracy $\uparrow$} & 
         \small{Precision $\uparrow$} & 
         \small{Recall $\uparrow$} & 
         \small{F1 $\uparrow$} & 
         \small{Accuracy $\uparrow$} & 
         \small{Precision $\uparrow$} & 
         \small{Recall $\uparrow$} & 
         \small{F1 $\uparrow$}
         \\
\midrule

    LLM
    & 75.39$\pm$3.32 & 75.66$\pm$3.44& 80.92$\pm$2.83 & 78.20$\pm$5.66
    & 74.38$\pm$6.68 & 78.66$\pm$5.31 & 75.16$\pm$4.32 & 76.87$\pm$5.65
    & 80.86$\pm$3.11 & 82.16$\pm$2.86 & 81.33$\pm$3.66 & 81.75$\pm$2.95 \\

    EANN & 71.53$\pm$0.91 & 71.38$\pm$1.23 & 63.82$\pm$2.11 & 68.91$\pm$1.58 & 70.17$\pm$0.79 & 71.28$\pm$1.32 & 67.36$\pm$2.17 & 69.10$\pm$1.83 & 79.18$\pm$0.76 & 80.31$\pm$1.23 & 78.52$\pm$0.32 & 79.44$\pm$2.13 \\
    SpotFake
    & 77.16$\pm$1.57 & 75.32$\pm$1.14 & 87.83$\pm$0.63 & 85.14$\pm$0.07 & 81.37$\pm$2.38 & 79.53$\pm$2.27 & 81.22$\pm$2.43 & 79.43$\pm$0.75 & 86.39$\pm$2.51 & 86.12$\pm$0.53 & 87.17$\pm$2.63 & 83.22$\pm$1.41 \\
    MVAE
    & 74.56$\pm$1.58 & 80.15$\pm$2.69 & 76.34$\pm$0.83 & 81.57$\pm$1.98 & 77.83$\pm$1.27 & 73.82$\pm$2.05 & 73.45$\pm$2.62 & 72.21$\pm$0.54 & 71.86$\pm$0.25 & 70.32$\pm$0.69 & 70.32$\pm$2.84 & 70.53$\pm$1.60 \\
    SAFE
    & 76.66$\pm$3.00 & 76.32$\pm$1.94 & 75.41$\pm$2.12 & 76.37$\pm$2.85 & 81.25$\pm$1.34 & 79.22$\pm$2.76 & 79.11$\pm$1.45 & 79.69$\pm$2.67 & 84.91$\pm$2.12 & 83.81$\pm$1.58 & 82.19$\pm$1.16 & 83.01$\pm$1.70 \\
    MCAN
    & 80.91$\pm$2.33 & 82.68$\pm$2.48 & 76.67$\pm$0.94 & 82.26$\pm$1.32 & 80.74$\pm$1.89 & 79.21$\pm$2.23 & 79.64$\pm$1.53 & 80.15$\pm$0.86 & 86.50$\pm$3.00 & 88.10$\pm$2.10 & 84.60$\pm$1.80 & 86.15$\pm$1.60 \\
    HMCAN
    & 83.91$\pm$1.49 & 81.68$\pm$2.08 & 84.67$\pm$1.21 & 82.57$\pm$1.62 & 86.36$\pm$1.83 & 83.18$\pm$1.41 & 83.81$\pm$2.51 & 83.49$\pm$1.07 & 86.75$\pm$2.95 & 88.40$\pm$3.00 & 84.65$\pm$1.80 & 87.20$\pm$1.20 \\
    FCN
    & 82.86$\pm$1.27 & 78.64$\pm$1.68 & 87.39$\pm$0.85 & 82.78$\pm$0.47 & 80.36$\pm$1.93 & 84.43$\pm$1.27 & 89.12$\pm$0.12 & 86.71$\pm$1.88 & 82.92$\pm$0.54 & 83.17$\pm$1.00 & 88.45$\pm$2.13 & 86.74$\pm$0.41 \\

 \cmidrule(lr){1-13}

    FCN-LP (HMCAN)
    & 84.57$\pm$1.62 & 83.58$\pm$1.66 & 85.22$\pm$2.06 & 84.04$\pm$0.82 & 87.25$\pm$1.18 & 84.48$\pm$1.91 & 84.78$\pm$1.95 & 84.50$\pm$0.85 & 87.15$\pm$1.32 & 88.82$\pm$2.56 & 86.55$\pm$1.98 & 88.11$\pm$1.66 \\
    FCN-LP (CLIP)
    & 85.32$\pm$2.56 & 81.52$\pm$2.82 & 89.32$\pm$0.99 & 85.24$\pm$1.93 & 84.68$\pm$0.81 & 86.32$\pm$1.55 & 89.85$\pm$1.22 & 87.97$\pm$0.88 & 84.47$\pm$1.66 & 88.41$\pm$0.26 & 91.18$\pm$0.69 & 89.78$\pm$0.84 \\

\cmidrule(lr){1-13}  

    FCN-LP (HMCAN) + LLM
    & 85.30$\pm$1.88 & 84.70$\pm$2.25 & 86.30$\pm$2.53 & 85.10$\pm$0.98 & 87.35$\pm$1.37 & 84.55$\pm$2.76 & 85.85$\pm$2.19 & 84.60$\pm$1.28 & 87.55$\pm$1.94 & 89.30$\pm$2.85 & 88.60$\pm$2.39 & 88.75$\pm$1.97 \\

    FCN-LP (CLIP) + LLM
    & 85.93$\pm$1.83 & 81.92$\pm$2.64 & 90.44$\pm$2.17 & 85.97$\pm$2.72 & 84.89$\pm$2.81 & 87.80$\pm$2.36 & 90.55$\pm$2.28 & 89.21$\pm$2.59 & 84.88$\pm$2.78 & 88.55$\pm$2.43 & 91.68$\pm$1.95 & 89.85$\pm$2.67 \\

\cmidrule(lr){1-13} 
    \textbf{GLPN-LLM (HMCAN)}
    & 87.60$\pm$1.23 & \textbf{86.52$\pm$0.92} & 88.88$\pm$1.88 & 86.86$\pm$0.85 & \textbf{88.29$\pm$0.38} & 86.92$\pm$0.88 & 88.14$\pm$1.11 & 86.87$\pm$1.32 & \textbf{90.66$\pm$1.32} & 89.06$\pm$0.13 & 92.20$\pm$0.17 & 91.46$\pm$0.16 \\

    \textbf{GLPN-LLM (CLIP)}
    & \textbf{88.83$\pm$2.23} & 84.02$\pm$2.43 & \textbf{92.68$\pm$2.81} & \textbf{89.03$\pm$2.18} & 86.47$\pm$1.97 & \textbf{89.24$\pm$2.74} & \textbf{92.13$\pm$2.56} & \textbf{90.66$\pm$2.32} & 86.74$\pm$1.32 & \textbf{89.83$\pm$0.88} & \textbf{93.27$\pm$0.65} & \textbf{91.52$\pm$0.66} \\
                        
\bottomrule        
    \end{tabular}
    }
        \caption{
        Performance comparison of different methods on the Twitter, PHEME, and Weibo datasets. The highest value in each column is marked in bold.
    }
    \label{tab:benchmark_maskllm}
\end{table*}

\section{Experiment}

In this section, we conduct experiments on public benchmark datasets to verify the effectiveness of GLPN-LLM and perform detailed analysis to assess the contribution of each proposed component.

\subsection{Datasets}

We evaluate our method on three widely-used benchmark datasets: Twitter~\cite{twitter}, PHEME~\cite{zubiaga2017exploiting}, and Weibo~\cite{jin2017multimodal}. The Twitter dataset contains 17,000 tweets (15,000 for training, 2,000 for testing). PHEME consists of 1,414 training tweets and 608 testing tweets related to five major news events. The Weibo dataset includes 4,141 training samples and 1,125 test samples from the Sina Weibo platform. 
Details of these datasets are provided in Appendix~\ref{sec:appendix-datasets}.

\subsection{Baselines}

We compare our GLPN-LLM framework against several state-of-the-art methods, including  LLM (GPT-4o)~\cite{hurst2024gpt}, EANN~\cite{wang2018eann}, SpotFake~\cite{spotfake}, and MVAE~\cite{mvae}, which leverage multimodal features for fake news detection. We also include SAFE~\cite{safe}, MCAN~\cite{mcan},  HMCAN~\cite{hmcan}, FCN~\cite{fcn-lp}, FCN-LP (HMCAN), FCN-LP (CLIP), FCN-LP (HMCAN) + LLM, and FCN-LP (CLIP) + LLM~\cite{fcn-lp}, where HMCAN and CLIP in parentheses denote the multimodal feature encoders.
FCN-LP + LLM is a naive solution, where FCN-LP directly uses LLM-generated pseudo labels.
For our method, we report the results of GLPN-LLM (HMCAN) and GLPN-LLM (CLIP), which use different multimodal encoders HMCAN and CLIP, respectively.
For experiments in ablation studies and detailed analysis, we use CLIP by default and omit the parentheses.

\subsection{Evaluation Metrics}
We use Accuracy, Precision, Recall, and F1 Score to evaluate detection performance. Following standard practice, F1 metrics are reported as macro averages, where higher values indicate better performance.

\subsection{Overall Evaluation}

The performance results on the three datasets are shown in Table~\ref{tab:benchmark_maskllm}. 
Based on the results, we have the following observations:
\begin{itemize}
\item Among the baselines that do not use label propagation techniques, HMCAN and FCN consistently outperform the others.
Note that LLM also shows poorer performance compared to these methods.
\item By incorporating label propagation techniques, FCN-LP achieves a noticeable improvement in performance across all datasets over baselines that do not use LP, which validates the efficacy of label propagation on this task.
\item FCN-LP + LLM leverages LLM-generated pseudo labels in a naive way, and only achieves marginal performance improvement over FCN-LP, showing that it is non-trivial to synergize LLMs with label propagation for fake news detection.
\item Our framework, GLPN-LLM, introduces a mask-based global label propagation module that works alongside an LLM-based pseudo label generation module, outperforming previous methods with substantial performance improvement.
Specifically, GLPN-LLM (HMCAN) demonstrates substantial improvements over FCN-LP (HMCAN) + LLM, and GLPN-LLM (CLIP) also shows considerable gains compared to FCN-LP (CLIP) + LLM. These consistent improvements are observed across all three datasets.
This shows that GLPN-LLM can effectively integrate LLM capabilities via label propagation techniques for multimodal fake news detection.

\end{itemize}

\begin{table*}[t]
    \centering
    \setlength{\tabcolsep}{4pt}
    
    \resizebox{1\linewidth}{!}{
    \begin{tabular}{lcccccccccccccc}
        \toprule
            \multirow{2}{*}[0pt]{\text{Method}}& 
            \multicolumn{4}{c}{{Twitter}} & \multicolumn{4}{c}{{PHEME}} &
            \multicolumn{4}{c}{{Weibo}} 
            \\
            \cmidrule(lr){2-5}
            \cmidrule(lr){6-9}
            \cmidrule(lr){10-13}
             &{Accuracy $\uparrow$} & 
             {Precision $\uparrow$} & 
             {Recall $\uparrow$}  & 
             {F1 $\uparrow$} & 
             {Accuracy $\uparrow$} & 
             {Precision $\uparrow$} & 
             {Recall $\uparrow$}  & 
             {F1 $\uparrow$} &
             {Accuracy $\uparrow$} & 
             {Precision $\uparrow$} & 
             {Recall $\uparrow$}  & 
             {F1 $\uparrow$}
             \\

\cmidrule(lr){1-13}

FCN-LP & 85.32 & 81.52 & 89.32 & 85.24 & 84.68 & 86.32 & 89.85 & 87.97 & 84.47 & 88.41 & 91.18 & 89.78 \\

GLPN & 85.47 & 82.27 & 90.57 & 86.30 & 85.58 & 87.58 & 89.99 & 86.96 & 85.07 & 88.95 & 91.61 & 90.76 \\

GLPN-LLM & \textbf{88.83} & \textbf{84.02} & \textbf{92.68} & \textbf{89.03} & \textbf{86.47} & \textbf{89.24} & \textbf{92.13} & \textbf{90.66} & \textbf{86.74} & \textbf{89.83} & \textbf{93.27} & \textbf{91.52} \\

\bottomrule        
    \end{tabular}   
    }
    \caption{
    Ablation study of different components of GLPN-LLM for multimodal fake news detection. The highest value in each column is marked in bold.
    }
    \label{tab:loss_ablation}
\end{table*}

\begin{figure*}[ht]
    \centering
    \includegraphics[width=1\linewidth]{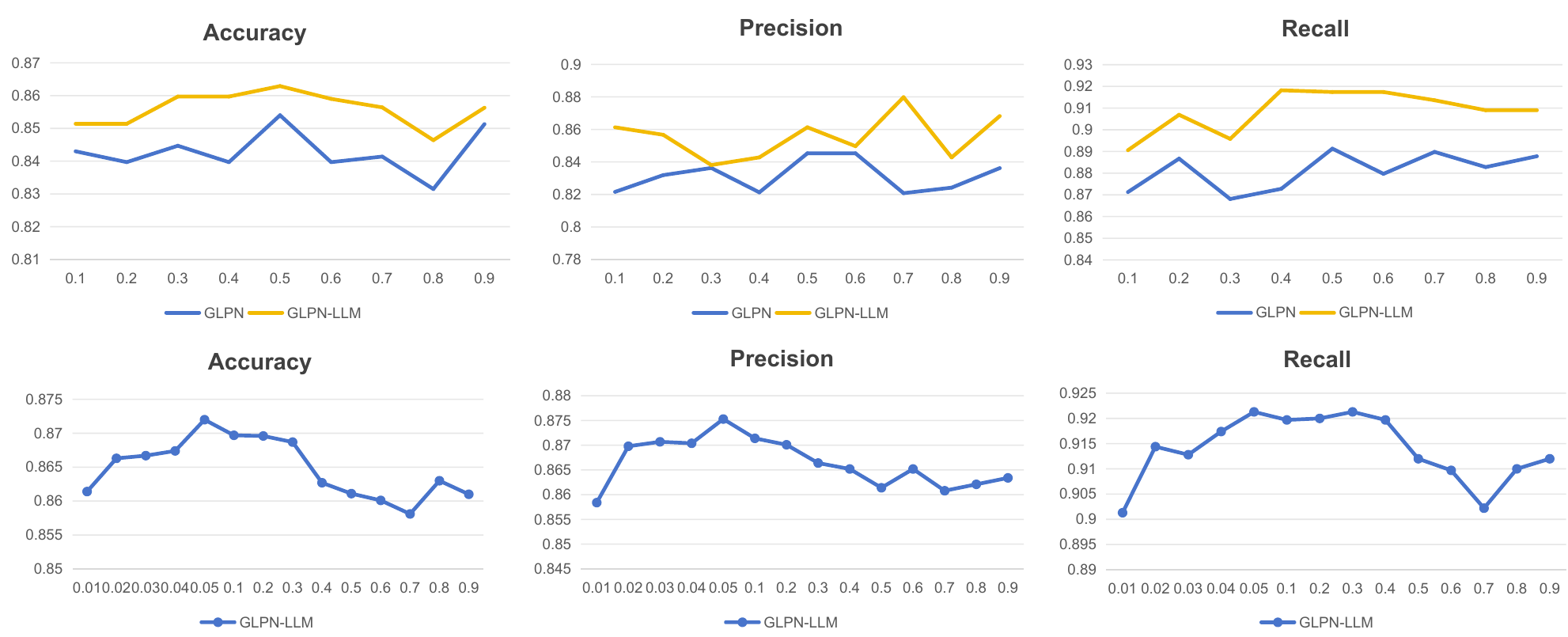}
    \caption{
    The effect of the mask rate and the impact of the quantity of LLM-generated pseudo labels.
    The first row shows how the mask rate parameter affects Accuracy, Precision, and Recall.
    The second row investigates how incorporating different proportion of LLM-generated pseudo labels from the test set into the label propagation process affects performance. The presented results are averaged across the three benchmark datasets.
        }
    \label{fig:curve}
\end{figure*}

\subsection{Ablation Study}

To assess the contribution of each component in our GLPN-LLM framework, we conduct an ablation study, summarized in Table~\ref{tab:loss_ablation}.
By introducing our mask-based global label-propagation module, GLPN surpasses FCN-LP, which utilizes basic label propagation techniques, demonstrating the superiority of our label propagation strategy.
GLPN-LLM extends GLPN by coupling our mask-based global label-propagation module with an LLM-based pseudo-label generation module, effectively integrating LLMs to achieve superior performance in multimodal fake news detection.

\subsection{Detailed Analysis}

\subsubsection{Effect of Mask Rate}
We examine the impact of the mask rate on the performance of GLPN-LLM, as shown in Figure~\ref{fig:curve}. The mask rate determines the proportion of label information that is masked during each training iteration, influencing how much label data is available for propagation. As illustrated in Figure~\ref{fig:curve}, we vary the mask rate from 0.1 to 0.9 and observe the corresponding changes in accuracy, precision, and recall. Our results indicate that a mask rate of 0.5 yields the best performance, with the model achieving an accuracy of 86.80\%, precision of 86.3\%, and recall of 91.8\%. This optimal mask rate suggests that balancing the availability and masking of label information is crucial for effective label propagation, allowing the model to generalize well without overfitting to specific label patterns.

\subsubsection{Impact of the Quantity of LLM-Generated Pseudo Labels}
We analyze how the number of LLM-generated pseudo labels affects detection performance, as shown in Figure~\ref{fig:curve}. We experiment with varying the percentage of pseudo labels integrated into the graph, ranging from the top 1\% to the top 90\% based on confidence scores. The results, depicted in Figure~\ref{fig:curve}, demonstrate that incorporating pseudo labels up to the top 5\% of confidence scores yields the highest performance improvements.
Increasing the percentage beyond 5\% does not lead to further gains and may even degrade performance due to the inclusion of lower-confidence labels, which can introduce noise into the label propagation process. Therefore, selecting a top 5\% threshold ensures that only high-confidence pseudo labels are utilized, enhancing the reliability and effectiveness of label propagation within the graph.

\subsubsection{Analysis of Prompt Designs}
\begin{figure}[!t]
    \includegraphics[width=1\linewidth]{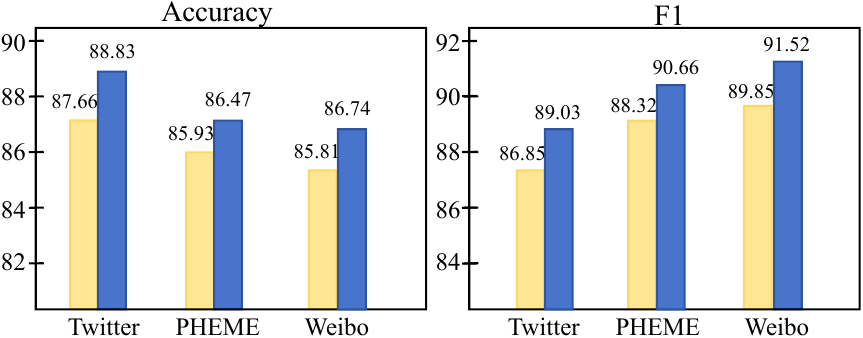}
    \caption{
    Effect of prompt detail on performance. Detailed prompts achieve higher accuracy and F1 scores than simple prompts. 
    Yellow bars represent simple prompts, while blue bars represent detailed prompts.
    }
    \label{fig:prompt-curve}
\end{figure}

We analyze the effect of prompt specificity by varying the level of contextual information provided, as shown in Table~\ref{table3} and Table~\ref{table4} in the appendix. In general, more detailed prompts tend to yield better performance as they provide clearer and more comprehensive guidance for the model. Prompts with sufficient detail ensure higher confidence scores and more reliable pseudo-label generation, resulting in improved recall rates and F1 scores, as shown in Figure~\ref{fig:prompt-curve}. In contrast, simple prompts, which lack necessary contextual information, often lead to poorer performance, lower confidence, and reduced label accuracy. However, it is important to strike a balance—while detailed prompts are beneficial, excessive complexity or overly intricate phrasing may introduce noise, potentially confusing the model and diminishing the effectiveness of label generation. Clear, concise, and well-structured prompts remain optimal for achieving consistent and reliable results.

\subsection{Case Study}

\begin{figure}[!t]
    \includegraphics[width=1\linewidth]{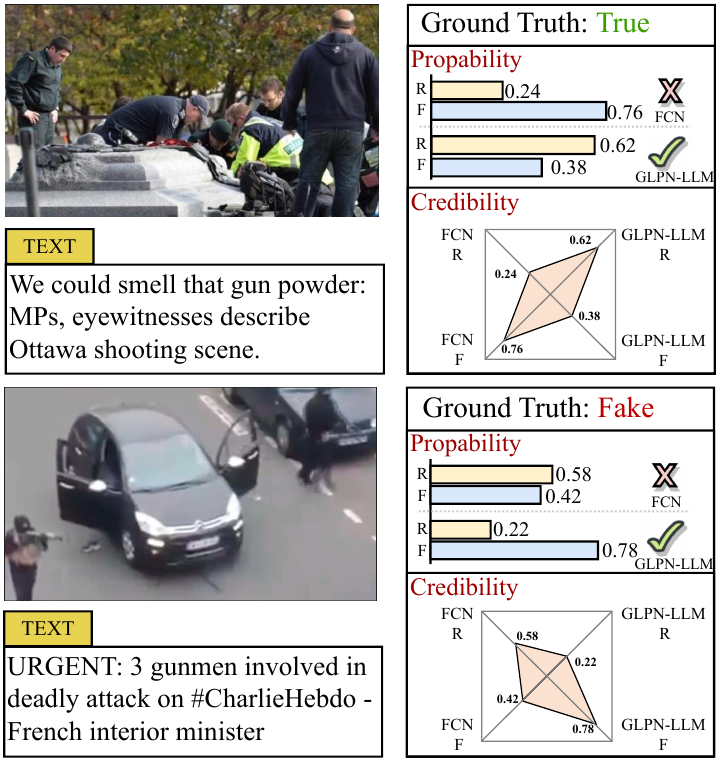}
    \caption{
    Case study - examples of challenging instances with their corresponding images and text.
    }
    \label{fig:casestudy}
\end{figure}

We present a case study demonstrating the classification process of our GLPN-LLM framework for real and fake news. As shown in Figure~\ref{fig:casestudy}, the model relies on textual content, image features extracted using CLIP, and LLM-generated pseudo labels. Real news is labeled `R', Fake news `F'. We report the model's predicted probability for the challenging instances.

In the first example, the original model misclassifies real news as fake. However, after applying the GLPN-LLM framework, the global label propagation module successfully corrects the classification, identifying the news as real. The model relied on the consistency of features across the dataset, ultimately leading to the correct classification despite initially low credibility scores for individual features. In the second example, the model initially misclassifies fake news as real. With the GLPN-LLM framework, the integration of text, image features, and LLM-generated pseudo labels helps the model accurately identify the news as fake. The label propagation mechanism ensures that the fake news label is correctly spread across related data points, rectifying the original misclassification. The model's decision is reinforced by the alignment between multiple features, resulting in the correct classification outcome.

\begin{figure}[ht]
    \centering
    \includegraphics[width=1\linewidth]{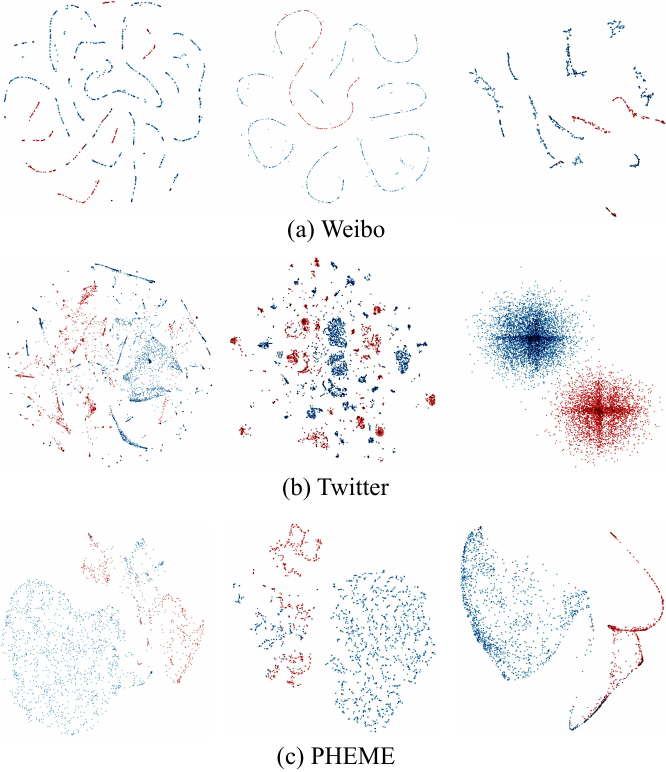}
    \caption{
    t-SNE visualizations of feature embeddings on the test set. The first column shows t-SNE embeddings from FCN, the second column shows embeddings from GLPN, and the third column shows embeddings from GLPN-LLM. Each point is color-coded according to its label.
    }
    \label{fig:tsne}
\end{figure}

\subsection{Visualization}

Figure~\ref{fig:tsne} presents t-SNE visualizations of the feature embeddings on the test set for three model configurations. 
The first column shows the embeddings produced by FCN; real and fake-news clusters largely overlap, indicating that the model struggles to distinguish the two classes. 
This overlap suggests that relying solely on FCN does not capture the nuanced differences between real and fake news. 
The second column depicts the embeddings from GLPN. With our mask-based global label propagation module, the separation between real and fake news becomes more pronounced—particularly on the Weibo dataset—demonstrating that GLPN yields embeddings in which the two classes are more distinct. 
The most substantial improvement appears when we integrate LLM-generated pseudo-labels (GLPN-LLM, third column). 
In this configuration the clusters are clearly separated, with minimal overlap, as illustrated on the Twitter dataset. 
This observation underscores the synergistic effect of combining label propagation with the rich semantic signals provided by LLM-generated labels.

\section{Conclusion}

In this paper, we present a framework, Global Label Propagation Network with LLM-based Pseudo Labeling (GLPN-LLM), for multimodal fake news detection.
While LLM-generated pseudo labels alone demonstrate poor performance compared to traditional detection methods, our approach effectively integrates LLM capabilities via label propagation techniques.
Experiments on three benchmark datasets demonstrate that GLPN-LLM consistently outperforms state-of-the-art baselines with significant improvements, highlighting the effectiveness of synergizing LLMs with label propagation for fake news detection.
In the future, our work will focus on exploring approaches to improve GLPN-LLM's scalability to larger and more complex datasets, while examining its adaptability across diverse social media platforms and content modalities to enhance practical applicability in real-world deployment scenarios.

\section*{Acknowledgement}
This work was supported in part by the National Natural Science Foundation of China under Grants 62206137, 62206200, 62276257, 62036012, in part by the Program for Young Talents of Science and Technology in Universities of Inner Mongolia Autonomous Region under Grant NJYT23105, and in part by the National Natural Science Foundation of Inner Mongolia under Grant 2025JQ012.

\section*{Limitation}

\noindent \textbf{Dependency on Backbone Models} 
The effectiveness of our GLPN-LLM framework is closely tied to the performance of the underlying backbone models, namely FCN. While these models provide strong feature representations, any limitations in their ability to capture comprehensive semantic relationships can directly impact the label propagation process. Consequently, the overall detection accuracy is highly dependent on the quality and robustness of these backbone models. Additionally, the reliance on specific backbones may limit the adaptability of our framework to other feature extraction architectures that might offer different advantages.

\noindent \textbf{Reliance on High-Confidence Pseudo Labels }
Our approach relies on the generation of high-confidence pseudo labels by the LLM to enhance label propagation. However, the accuracy of these pseudo labels is contingent upon the LLM's ability to produce reliable predictions. Inaccurate or biased pseudo labels can introduce noise into the label propagation process, potentially degrading the model's performance. Ensuring the reliability of pseudo labels is crucial, and future work may explore more robust methods for pseudo label verification and refinement to mitigate this limitation.

\bibliography{custom}

\appendix

\section{Datasets}
\label{sec:appendix-datasets}
We validate the proposed methods on three real social media datasets: Twitter~\cite{twitter}, PHEME~\cite{zubiaga2017exploiting}, and Weibo~\cite{jin2017multimodal}:
\\
\noindent \textbf{Twitter}~\cite{twitter} is used to detect fake content on social media by analyzing approximately 17,000 unique tweets.
The tweets are originally from the widely-used fake news dataset MediaEval Verifying Multimedia Use benchmark~\cite{twitter}.
Each tweet includes textual content and a related image, with labels for real/fake.
Following the benchmark~\cite{fcn-lp}, we split the dataset into train and test sets with 15,000 and 2,000 tweets, respectively.
\\
\noindent\textbf{PHEME}~\cite{zubiaga2017exploiting} includes tweets sourced from Twitter, specifically targeting five significant breaking news events. 
Each event includes a substantial collection of tweets, along with their textual content, associated images, and real/fake labels.
Following the setup in the benchmark~\cite{fcn-lp}, we adopt 1,414 and 608 tweets as the training and test sets, respectively. 
\\
\noindent\textbf{Weibo}~\cite{jin2017multimodal} originates from Sina Weibo, a widely used microblogging platform in China.
Following the setup in the benchmark~\cite{fcn-lp}, we adopt 4,141 and 1125 tweets as the training and test sets, respectively. 

\section{Baselines}
\label{sec:appendix-baseline}
We benchmark our proposed GLPN-LLM framework against several state-of-the-art methods that utilize multimodal features and label propagation techniques. The combined effectiveness of the backbone and label propagation is significantly influenced by the backbone's ability to extract and integrate features, with weaker backbones limiting overall performance. Building upon recent advancements in label propagation~\cite{fcn-lp}, we include only methods that employ state-of-the-art backbone architectures.

We provide a comprehensive comparison with leading multimodal and label propagation approaches.
\begin{itemize}
    \item EANN~\cite{wang2018eann} employs attention mechanisms to integrate textual and visual features for fake news detection, focusing on attention-based fusion of modalities to enhance prediction accuracy.
    \item SpotFake~\cite{spotfake} utilizes multimodal information to identify fake news by analyzing both textual content and accompanying images, optimizing feature alignment across text and images for effective detection.
    \item MVAE~\cite{mvae} is a multimodal variational autoencoder that captures the joint distribution of text and image data, improving fake news classification through the combination of textual and visual information.
    \item MCAN~\cite{mcan} is a multi-modal contextual attention network that fuses inter-modality and intra-modality relationships, enhancing fake news detection by modeling the dependencies between text and image modalities.
    \item SAFE~\cite{safe} combines multimodal feature extraction with cross-modal similarity measures to learn tweet representations, directly measuring the similarity across modalities to achieve effective alignment for detecting fake news.
    \item HMCAN~\cite{hmcan} utilizes a hierarchical multi-modal contextual attention network to capture rich hierarchical semantics, enhancing fake news detection by modeling both high-level and fine-grained relationships in the data.
    \item FCN~\cite{fcn-lp} utilizes CLIP for feature extraction. It constructs a cross-modal tweet graph to unify text and image features and then employs a Graph Convolutional Network (GCN) for classification.
    \item FCN-LP~\cite{fcn-lp} uses CLIP for feature extraction, followed by fixed label propagation. It builds a cross-modal tweet graph to unify text and image features, utilizing iterative label propagation to refine predictions.
\end{itemize}

To ensure fairness in our comparisons, we follow the benchmark setup of FCN-LP~\cite{fcn-lp} and use the same similarity threshold \( \theta \) across all datasets to construct the cross-modal tweet graph.

\section{Implementation Details}
For our GLPN-LLM framework, the core graph neural network is a Graph Convolutional Network (GCN)~\cite{kipf2016semi}.
We adopt Adam~\cite{adam} as the optimizer with a 1e-3 learning rate.
The hyperparameters for label propagation mask rate $\lambda_{\text{mask}}$ and LLM pseudo rate $\lambda_{\text{pseudo}}$ are set to 0.50 and 0.05, respectively.
The multimodal common space dimension learned by GCN is set to 512.
To enable a consistent comparison with the baseline, we follow the settings of previous work  FCN-LP~\cite{fcn-lp} and set the similarity threshold to \(\theta = 0.95\) for the Twitter, PHEME and Weibo datasets when building the cross-modal tweet graph.
We train the model 5 times and report the average and standard deviation for accuracy, precision, recall, and F1 score.

\section{Efficiency}
\label{sec:efficiency}

The LLM module operates independently of the Label Propagation module and utilizes an API to significantly reduce computational overhead. Its computational complexity is $O(N_{\text{test}} \cdot T_{\text{avg}})$, where $N_{\text{test}}$ is the size of the test set and $T_{\text{avg}}$ is the average number of tokens per sample (input + output). Taking the Twitter dataset as an example, the LLM exclusively processes tweets from the test set. Based on GPT-4o API rates, the processing cost per tweet is approximately \$0.00074, which renders the module highly efficient for deployment. This high efficiency ensures that the module is both cost-effective and scalable for practical deployment.

The LP module introduces minimal additional cost. This is achieved by injecting pseudo labels (converted to one-hot vectors) into node features and then propagating them across the graph. Its complexity for $K$ iterations on a graph with $M$ edges is $O(K \cdot M)$. Consequently, the label propagation step increases the overall computation time by only about 2\%, while substantially boosting performance.

Overall, the framework's computational costs, both in time and monetary terms, remain low and are more than justified by the resulting accuracy improvements, making the approach practical for real-world scenarios.

\begin{table*}[h!]
    \centering
    \caption{Prompts for Mixed-Initiative Labeling}
    \label{tab:prompts_labeling}
    \renewcommand{\arraystretch}{1.3} %
    \begin{tabular}{p{2.5cm} p{13cm}} %
        \hline
        \textbf{Role} & \textbf{Content} \\ \hline
        system & You are a professional misinformation evaluation expert with extensive experience in detecting and evaluating fake news. 
        Your primary task is to assess the authenticity of the provided news content. 

        You must adhere to the following strict evaluation guidelines:
        \begin{itemize}
            \item If the news is clearly true, label it as 1 (true).
            \item If the news contains ambiguity, unverifiable information, or suspicious claims, you must classify it as 0 (false).
            \item Alongside your classification, provide a confidence score (between 0\% and 100\%) that reflects your certainty in the decision.
            \item Your confidence score should be lower (e.g., 50\%-70\%) when the news includes unclear or mixed signals, and higher (e.g., 80\%-100\%) when you are certain.
        \end{itemize}

        Output format:  
        \texttt{Result: R, Confidence: C\%,}  
        where \texttt{R} is 1 (true) or 0 (false), and \texttt{C} is the confidence score. Be precise, concise, avoid unnecessary explanations, and give me the reason. \\ \hline
        
        user & BREAKING: SkyBusiness reports another five hostages seen escaping \#sydneysiege. \newline
        Additional context: The event is unfolding in Sydney's central business district. Initial unverified reports mention hostages being rescued, but official statements have not yet been released. \\ \hline
        
        system & Result: 1, Confidence: 85\% \newline
        Reason: Based on credible news reports and consistent information across major media outlets, the claim of hostages escaping appears highly plausible. Minor uncertainty remains due to the absence of official verification. \\ \hline

        user & CONFIRMED: NASA discovers alien life on Mars. \\ \hline
        
        system & Result: 0, Confidence: 30\% \newline
        Reason: This claim lacks supporting evidence from verified scientific sources, and NASA has not released any official confirmation regarding such a discovery. The headline seems sensationalized or misleading. \\ \hline

        user & ALERT: Severe storms expected to hit California tomorrow, warns National Weather Service. \\ \hline
        
        system & Result: 1, Confidence: 95\% \newline
        Reason: The information originates from the National Weather Service, a highly reliable and authoritative source. Severe weather forecasts for tomorrow are consistent across official meteorological channels. \\ \hline
    \end{tabular}
    \label{table3}
\end{table*}

\begin{table*}[h!]
    \centering
    \caption{Simple Prompt for Mixed-Initiative Labeling}
    \label{tab:simple_prompt}
    \renewcommand{\arraystretch}{1.3} %
    \begin{tabular}{p{2.5cm} p{13cm}} %
        \hline
        \textbf{Role} & \textbf{Content} \\ \hline
        system & You are tasked with determining whether the provided news content is true or false.  
        \newline Output format: \texttt{Result: R, Confidence :c}, where \texttt{R} is 1 (true) or 0 (false). \\ \hline

        user & BREAKING: SkyBusiness reports another five hostages seen escaping \#sydneysiege. \\ \hline

        system & Result: 1, Confidence: 49\% \\ \hline

        user & CONFIRMED: NASA discovers alien life on Mars. \\ \hline

        system & Result: 0, Confidence: 20\% \\ \hline

        user & ALERT: Severe storms expected to hit California tomorrow. \\ \hline

        system & Result: 1, Confidence: 63\% \\ \hline
    \end{tabular}
    \label{table4}
\end{table*}

\end{document}